\documentclass[10pt,twocolumn,letterpaper]{article}

\usepackage[pagenumbers]{cvpr} 

\usepackage[dvipsnames]{xcolor}

\newcommand\embed[0]{\boldsymbol{\ell}}

\usepackage[pagebackref,breaklinks,colorlinks]{hyperref}

\usepackage[misc]{ifsym}

\usepackage{amsmath}
\usepackage{caption}
\usepackage{colortbl}
\usepackage{mathalfa}

\usepackage{threeparttable}
\usepackage{mathrsfs}

\usepackage{booktabs}

\usepackage{graphicx}
\usepackage{caption}  

\numberwithin{equation}{section}
\def\confName{3DV\xspace}
\def\confYear{2024\xspace}

\title{Fast High Dynamic Range Radiance Fields for Dynamic Scenes}

\author{%
  Guanjun Wu$^{1\star}$, \quad Taoran Yi$^{2\star}$, \quad Jiemin Fang$^{2^\dag}$, \quad Wenyu Liu$^{2}$, 
  \quad Xinggang Wang$^{2^{\dag\textrm{\Letter}}}$\\
  $^1$School of CS, Huazhong University of Science and Technology\\
  $^2$School of EIC, Huazhong University of Science and Technology\\
  \texttt{\small\{guajuwu, taoranyi, liuwy, xgwang\}@hust.edu.cn} \; \texttt{\small jaminfong@gmail.com} 
}

\begin{document}

\twocolumn[
{%
\renewcommand\twocolumn[1][]{#1}%
\maketitle
\vspace{-20pt}
\begin{center}
\centering

\includegraphics[width=1.0\linewidth]{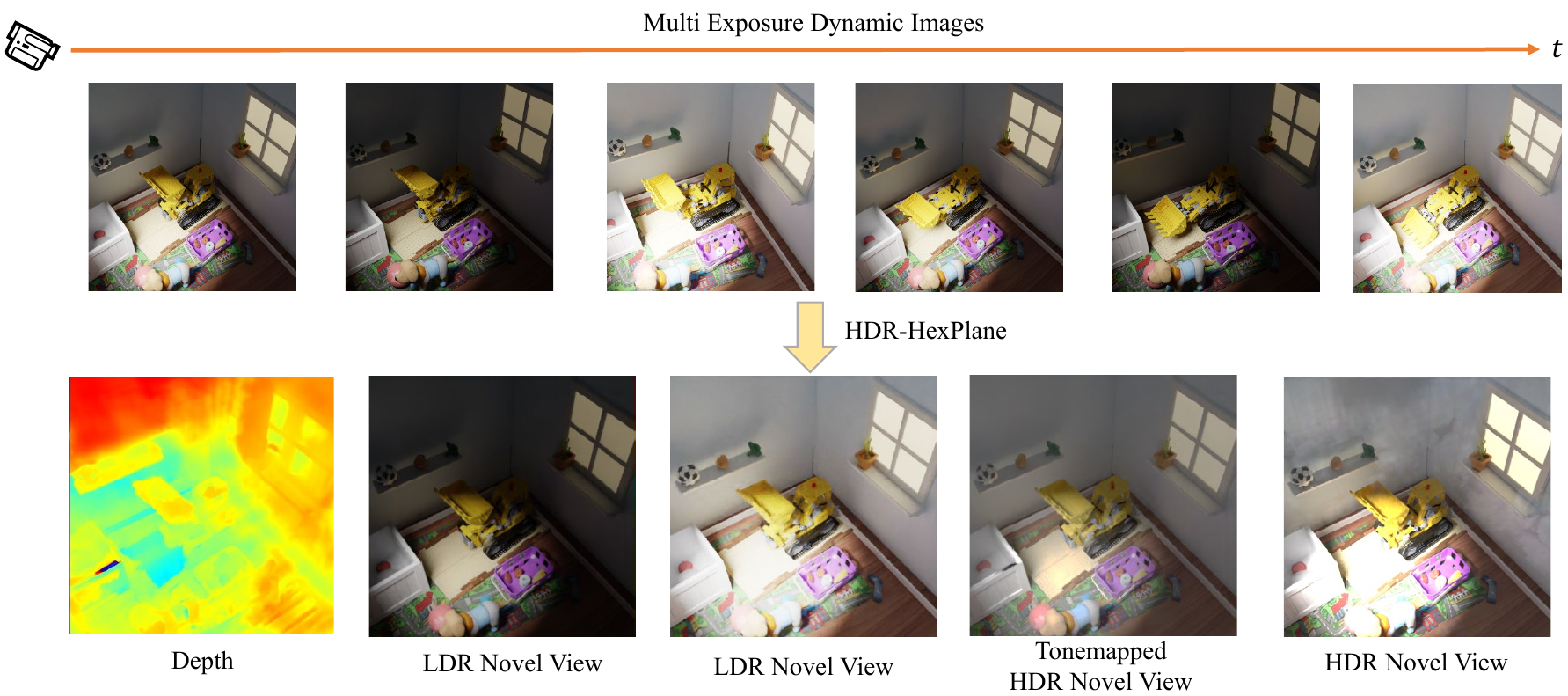}
\vspace{-20pt}
\captionof{figure}{Our model is capable of synthesizing novel viewpoint images in dynamic scenes by capturing images with different exposure values at different time points. Additionally, it can seamlessly combine images with varying exposures and produce high dynamic range (HDR) images. With the tone mapping function applied, a better color balance is achieved, enhancing the overall visual quality of images.
}
\label{fig:mutiexposure}
\end{center}%
}
]

\let\thefootnote\relax\footnotetext{$^\star$Equal contributions; $^\dag$Project lead; $^{\textrm{\Letter}}$Corresponding author.}

\begin{abstract}
Neural Radiances Fields (NeRF) and their extensions have shown great success in representing 3D scenes and synthesizing novel-view images. However, most NeRF methods take in low-dynamic-range (LDR) images, which may lose details, especially with nonuniform illumination. Some previous NeRF methods attempt to introduce high-dynamic-range (HDR) techniques but mainly target static scenes. To extend HDR NeRF methods to wider applications, we propose a dynamic HDR NeRF framework, named HDR-HexPlane, which can learn 3D scenes from dynamic 2D images captured with various exposures. A learnable exposure mapping function is constructed to obtain adaptive exposure values for each image. Based on the monotonically increasing prior, a camera response function is designed for stable learning. With the proposed model, high-quality novel-view images at any time point can be rendered with any desired exposure. We further construct a dataset containing multiple dynamic scenes captured with diverse exposures for evaluation. All the datasets and code are available at \url{https://guanjunwu.github.io/HDR-HexPlane/}.
\end{abstract}

\section{Introduction}{\label{sec:intro}}
Novel view synthesis has been a hot topic in the field of 3D vision. With a set of 2D images as input, a new image needs to be generated from a novel view that maintains geometric consistency with the other images. 
NeRF (Neural Radiance Fields)~\cite{nerf} proposes to model the scene with an implicit function, \eg MLP~\cite{mlp}, and produces high-quality novel-view images.
Over the years, the research community has significantly expanded the capabilities of NeRF: accelerating the training and rendering speed~\cite{dvgo,tensorf,instantngp}, extending to dynamic scenes~\cite{dnerf,dynnerf,tineuvox,hexplane,lin2023im4d,xu20234k4d,nerfplayer}, learning from unposed images~\cite{inerf,nerf--,barf,garf,nopenerf,localrf,l2gnerf} \etc.

However, NeRF training requires input images to present accurate and complete scene information. Overexposed or underexposed regions are not expected whose colors may significantly diverge from the intrinsic colors, leading to severe cross-view inconsistency and damaging the learning process. The HDR (High Dynamic Range)~\cite{recoveringhdr} technology addresses the issue by capturing images with varying brightness levels. Taking images with different exposures as input, HDR image synthesis can produce an image containing regions of varying brightness and darkness. Several works~\cite{hdrnerf,hdrplenoxels,casualindoorHDR} have attempted to incorporate the HDR technology into the NeRF framework and achieved high rendering quality. However, these existing methods mainly rely on images captured with varying exposure from static scenes, which may not meet the requirements for wider applications.

Real-world scenarios usually involve dynamic scenes, where complex point motions or structure changes exist. Representing dynamic scenes with a high dynamic range is essential. Examples of scenes with strong light sources include indoor scenes with bright lighting, outdoor scenes under intense sunlight, scenes illuminated by flames, and scenes with reflections and mirror-like surfaces. In these scenes, the presence of strong light sources leads to high contrast between light and dark areas, resulting in high dynamic range images. Additionally, objects in these scenes are likely to move from darker to brighter areas. Therefore, the model should be able to learn accurate scene representations from inputs of dynamic scene images captured at different exposures and multiple viewpoints. Thus making the NeRF model available for handling HDR images will make it possible to learn accurate scene structure even under nonuniform illumination.

Based on the fast training HexPlane~\cite{hexplane} representation, we propose HDR-HexPlane, a neural radiance field method for dynamic scenes where input images are captured with various exposures, in which an image exposure learning module is proposed to automatically obtain the exposure coefficient for each image. 
Additionally, due to the instability of exposure values and camera response curves, we instantiate the camera response function as a common fixed curve, which provides monotonically increasing priors and eases the optimization phase.
To benchmark the HDR novel view synthesis problem for dynamic scenes, we construct a new dataset captured with both single-camera and multi-camera setups with variable exposure parameters. These scenes contain deformable objects and diverse brightness due to changing exposures and nonuniform illumination.

Our contributions can be summarized as follows.
\begin{itemize}
    \item  We propose an end-to-end NeRF framework, \ie HDR-HexPlane, for high-dynamic-range dynamic scene representation, allowing for efficient scene learning and novel view synthesis based on images captured at different exposure levels. HDR-HexPlane enjoys the flexibility to adjust the exposure levels as desired and the ability to render a balanced image considering both over/under-exposed regions.
    \item An adaptive algorithm is proposed for efficiently and accurately learning the exposures of each captured image, freeing the requirement process of camera exposure parameters.
    \item We contribute a dataset containing dynamic scenes captured under both single-camera and multi-camera setups, with various exposure values. This dataset serves as an evaluation benchmark for view synthesis on HDR dynamic scenes.
\end{itemize}
\section{Related Works}

\subsection{Dynamic Neural Radiance Fields}
Initially, methods like~\cite{bakingnerf,neusample} attempt to accelerate the rendering process of NeRF, which are known for being slow. Subsequently,~\cite{dvgo,tensorf,instantngp} introduce voxel grids as an explicit scene representation, significantly speeding up the training process and reducing NeRF's training time from days to minutes.~\cite{dnerf,dynnerf} break the assumption of static scenes and extend the new view synthesis based on NeRF to dynamic scenes.~\cite{tineuvox} utilizes explicit representations to accelerate the training and rendering of dynamic scenes.~\cite{mononerf,gneuvox} focus on the generalization problem in dynamic scenes.~\cite{hexplane,kplanes,dtensorf} employ low-dimensional plane grids as substitutes for high-dimensional space grids, effectively reducing the spatial complexity of the scene and improving training efficiency.

However, these models are based on input images that are captured in low dynamic ranges. If there is a significant contrast in brightness within the scene, the captured data may suffer from overexposure or underexposure, making it challenging to effectively model dynamic scenes. In our work, we extend the capabilities of efficient dynamic scene learning, allowing for the modeling of dynamic scenes under various shooting conditions.

\subsection{HDR Imaging}
The classic HDR algorithm~\cite{recoveringhdr} is based on a set of scene images captured at different exposures and their corresponding exposure values. It uses the least squares method to estimate the camera response curve and generate the HDR images.

Deep learning-based approaches~\cite{deepfuse,hdrGAN} can synthesize high-resolution HDR images by training on a dataset of HDR ground truth images.~\cite{zerodce} enhances images using self-supervised learning. It can generate comparable HDR-like images from single under-exposed or over-exposed images. Recently,~\cite{enhance1,enhance2,enhance3,enhance4,imageenhancer,wang2023glowgan} success in low-light enhancement or LDR2HDR imaging. Though high-quality images can be synthesized, the novel view synthesis is ignored.

\cite{deephdrfordynamicscene} is capable of synthesizing HDR images from three different exposed images of dynamic scenes. \cite{flownet,flownet2.0,deephdrdynamicscene} utilize optical flow to synthesize HDR images, achieving good results. However, these methods are not end-to-end and have limited functionality, making it difficult to effectively control the temporal information of synthesized images.

To address these limitations, we introduce the dynamic neural radiance field into HDR image synthesis. This approach effectively solves the aforementioned issues by allowing the synthesis of images from new perspectives while preserving temporal coherence.

\subsection{HDR Novel View Synthesis}
The training of NeRF~\cite{nerf} is based on the assumption of multi-view consistency. However, guaranteeing simultaneous multi-view images captured at a particular moment during data acquisition is challenging. As time progresses, changes in environmental lighting conditions occur, which violate the assumption of multi-view consistency.~\cite{nerfinthewild} addresses this issue by introducing appearance embedding, which learns different appearances for each image. This approach alleviates the problem above and has been applied in~\cite{meganerf,blocknerf}.

On the other hand,~\cite{hdrplenoxels,hdrnerf} modify the exposure settings during image capture to handle the low dynamic range of camera-captured images. By doing so, these methods enable the models to learn representations of high dynamic range scenes, accommodating both overexposed and underexposed regions.~\cite{casualindoorHDR} uses LDR2HDR image enhancement network to generate HDR images as prior for training neural radiance fields, which also succeeds in rendering HDR novel view.~\cite{adop} which requires dense colmap sfm results, uses the point-cloud representation to render HDR novel views. 

However,~\cite{nerfinthewild} cannot explicitly edit the scene's exposure and render HDR novel views.~\cite{hdrplenoxels,hdrnerf,casualindoorHDR,adop} are limited to static scenes. Our proposed method extends its boundary into HDR dynamic novel view rendering with correct depth value like Fig.~\ref{fig:mutiexposure}. In addition, it is worth noting that currently there are no available datasets for evaluating the model. Therefore, we provide a new dataset for testing the synthesis of new viewpoints in dynamic HDR scenes.
\begin{figure*}[th]
  \centering
    \vspace{-10pt}
\includegraphics[width=\linewidth]{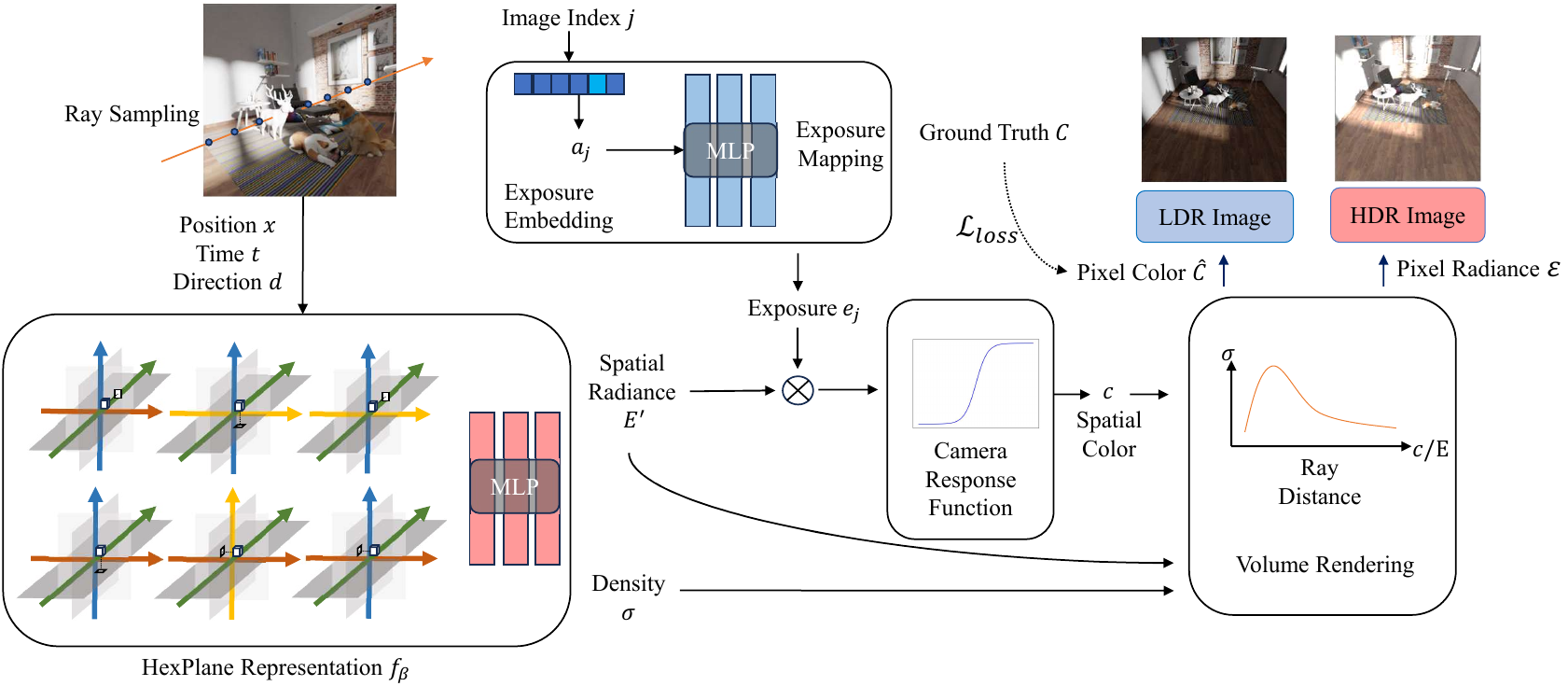}
    \vspace{-20pt}
    \caption{The overall framework of our proposed method. Firstly, we cast multiple rays from the camera and sample a series of points $x$ from each ray. These points, along with the current timestamp $t$ and direction $d$, are fed into the HexPlane module. HexPlane calculates the radiance value $E^{\prime}$ and density $\sigma$ corresponding to each point, which allows us to render HDR images using the volume rendering equation. Simultaneously, the exposure mapping module learns the corresponding logarithmic exposure coefficients $e_j$ for each image index $j$. After multiplying $E^{\prime}$ with the calculated color $c$ using the camera response function, we render the corresponding LDR image using the volume rendering equation.}
    \vspace{-25pt}
    \label{fig:pipeline}
  \hfill
\end{figure*}

\section{Methods}
In this section, we simply review the formulations of HDR-NeRF~\cite{hdrnerf} and HexPlane~\cite{hexplane} in Sec.~\ref{preliminary}. Then we introduce the framework of HDR-HexPlane in Sec.~\ref{hdr training pipeline}. We discuss how to learn the unknown exposure in Sec.~\ref{learn-exp-ebd}, and the sigmoid camera response function is introduced in Sec.~\ref{sigmoid CRF}. Finally, the optimization part is discussed in Sec.~\ref{Sec_Optimization}.

\subsection{Preliminary}\label{preliminary}

\paragraph{HDR-NeRF.}\label{hdr-nerf}

NeRF~\cite{nerf} enables high-quality synthesis of novel views through implicit scene representation and the volume rendering equation~\cite{volumerendering}. HDR-NeRF~\cite{hdrnerf}, on the other hand, introduces camera response function learning, tone-mapping, and incorporates exposure time of the scene, allowing NeRF to take images with different exposure values as input for training and perform HDR viewpoint synthesis. For each ray, $r = o+ t d$, HDR-NeRF and NeRF follow a similar process by sampling a series of points $\{x_1,x_2,...,x_n\}$ along the ray. For each point $x$ and direction $d$, HDR-NeRF employs an MLP network $f_\theta$~\cite{mlp} to compute logarithmic space radiance value $E^{\prime} = \ln{E(r)}$ and volume density $\sigma(r)$.

\begin{equation}\label{nerf}
    (E^{\prime}, \sigma) = f_\theta(x, d).
\end{equation}

Then, utilizing the method proposed by \cite{recoveringhdr}, incorporating the logarithmic value of the exposure $e_j$, and finally applying an MLP to fit the camera response function $g$, the final spatial color value $c$ is obtained.
\begin{equation}
c_{j}(E,e_j) = g(\ln{E} + \ln{e_j}).
\end{equation}\label{hdr-pipeline}
The volume rendering equation~\cite{volumerendering} is used to aggregate the spatial color value $c$ and volume density $\sigma$ of each point along a ray to obtain the pixel color $\hat{C}$.
\begin{equation}
    \begin{aligned}
        &\ \hat{C}(r) = \int_{t_n}^{t_f}T(t)\sigma(r(t))c(r(t),d)dt\ \\
&\ where\ T(t)=\exp{(-\int_{t_n}^t\sigma(r(s))ds}.\\
    \end{aligned}
    \label{volume rendering}
\end{equation}

In~\cite{hdrnerf}, color $c$ and exposure $e_i$ are known, while radiance $E_i$ and camera response curve $g$ are unknown. In this situation, scaling $E$ by $\alpha$ and giving $\ln{\alpha}$ offset to $g$ may result in the same outcome. Therefore, both~\cite{recoveringhdr,hdrnerf} use the zero-point constraint to fix $g(0)$ to a constant $C_0$, which gives the camera response function $g$ a proper prior.
\begin{equation}
    \begin{aligned}
        \mathcal{L}_u=\left\|g(0)-C_0\right\|.
    \end{aligned}
\end{equation}\label{eq:zeropointconstraint}
The total loss is defined summantion of reconstructed loss $\mathcal{L}_c$ and zero-point constraint loss $\mathcal{L}_u$:
\begin{equation}
    \begin{aligned}
        \mathcal{L} = \mathcal{L}_c + \lambda_u \mathcal{L}_u.
    \end{aligned}
\end{equation}
\paragraph{HexPlane.} \label{hexplane}
HexPlane \cite{hexplane} is an effective representation of dynamic scene reconstruction. Based on~\cite{tensorf}, it combines both temporal and spatial information into 6 learnable parametric planes:
\begin{equation}
\begin{aligned}
        D &= \sum_{r=1}^{R_1}M_r^{XY} \odot M_r^{ZT} \odot v_r^1 + \sum_{r=1}^{R_2}M_r^{XZ} \odot M_r^{TY} \odot v_r^2 \\
    &+ \sum_{r=1}^{R_3} M_r^{YZ} \odot M_r^{XT} \odot v_r^3, 
\end{aligned}
\end{equation}
where $M_r^{AB} \in \mathbb{R}^{AB}$ is a plane of features, and $D$ stands for the hexplane's voxel grid representation.

Given the position $x$ and time $\mathbf{t}$,~\cite{hexplane} utilizes them as the query vector to compute hidden information with six bilinear interpolations and a vector-matrix product. Meanwhile, positional encoding $\gamma$ is applied to position $x$, time $\mathbf{t}$, view direction $d$ into high dimension variable; Then all the variables are concatenated together and fed into separate MLPs to export the space color $c$ and density $\sigma$. Finally, the volume rendering equation Eq.~(\ref{volume rendering}) is employed to compute the integrated color.

\subsection{Overall Framework}\label{hdr training pipeline}
Since the input images are LDR images captured at different exposures, the geometric information of the scene remains unchanged with exposure variations. Similar with~\cite{hdrnerf}, during training, for each corresponding ray's origin $o$ , direction $d$ and time $\mathbf{t}$, we 
 sample the points $x$ on the ray between distances $near$ and $far$, then compute the volume density $\sigma$ and logarithmic space radiance value $E^{\prime} = \ln{(E)}$ using HexPlane~\cite{hexplane} representation $f_\beta$:
\begin{equation}
(E^\prime,\sigma) = f_{\beta}(x,d,\mathbf{t}),    
\end{equation}
Then we can compute the spatial color of the point $c_{j}$ with function $g$ by logarithmic space radiance value $E^\prime$ and logarithmic exposure coefficient $e^\prime_j = \ln{(e_j)}$ :
\begin{equation}
    c_{j} = g(\ln{E} + \ln{e_j}), \label{hdrcolor}
\end{equation}
while the exposure depends on index of the image $e_j = \phi(j)$, so the Eq.(~\ref{hdrcolor}) can be written as :
\begin{equation}
    c_{j} = g(\ln{E} + \ln{\phi(j)}) .\label{hdrcolor_image}
\end{equation}

And we can use the volume rendering equation Eq.~(\ref{volume rendering}) to compute the final result. The spatial color $c$ and density $\sigma$ are combined to obtain the pixel color $\hat{C}$, while the spatial radiance $E$ and density $\sigma$ are used to compute the pixel radiance $\mathcal{E}$. Then the parametric function $\phi$ is discussed in Sec.~\ref{learn-exp-ebd}. The whole framework is depicted in Fig.~\ref{fig:pipeline}. The module we designed explicitly separates the modeling of dynamic scenes and the learning of scene illumination, thereby addressing the issue of inconsistent colors in multi-view dynamic scenes caused by objects moving from dark to bright regions. Specifically, we let HexPlane learn the dynamic scene independently of exposure and only predict its spatial radiance value $E$ and volume density $\sigma$. The exposure learning and camera response function module then maps the space radiance values output by HexPlane to corresponding LDR values for different exposures, combined with the exposure-independent volume density $\sigma$, thus rendering LDR images with geometric consistency.

\subsection{Exposure Mapping}\label{learn-exp-ebd}

Following the traditional HDR image recovering pipeline 
 \cite{recoveringhdr}, we also map the spatial radiance and exposure into the logarithm domain. The Eq.~(\ref{hdrcolor}) can be written as :
\begin{equation}
    c_{j}= g(\ln{E} + \ln{e_j}).
\end{equation}
Certain points may be redundantly selected when performing ray sampling from different viewpoints. These sampled points are utilized in computing their corresponding spatial colors following the volume rendering equation (Eq.~\ref{volume rendering}). As a result, for ground truth images captured by the same camera settings with different exposure values, color consistency still exists. Consequently, the exposure values corresponding to these images can be jointly optimized. This observation leads us to believe that it is indeed possible to learn the corresponding exposure coefficients. 
Specifically, for each image index $j$, we assign it a feature embedding $a_j = {\embed}(j)$ and then utilize an exposure MLP $\phi_e$ to compute its exposure $e_j$.
\begin{equation}
\begin{aligned}
    e_j = \phi_e({\embed}(j)).
\end{aligned}
\end{equation}

Using the exposure MLP ensures the optimization of exposure embeddings more smoothly and facilitates convergence. In our pipeline, we designate the camera response function $g$ as fixed and set it to a sigmoid function, as discussed in Sec.~\ref{sigmoid CRF}. Instead, we use trainable parameters to learn the exposure values $e_i$. 

\subsection{Camera Response Function}\label{sigmoid CRF}

HDR-NeRF~\cite{hdrnerf} proposes a trainable camera response function (CRF) with known exposure values. When exposure values are unknown, estimating CRF and exposure values may become difficult (See in Sec.~\ref{sec_ablation}.) To solve the issue, we fix the CRF as a known function which should be: 1) Monotonically increasing and smooth, and 2) having upper and lower bounds to limit its range from 0 to 1.

Therefore, we consider using the formula of the sigmoid function:
\begin{equation}
f(x) = \frac{1}{1+e^{-x}}.
\end{equation}
That means when training the HDR-HexPlane, we enable model to learn the exposure value across input LDR images under the certain camera settings and the learned exposure value among the different scenes are also comparable. What's more, explicitly determining the formula of the camera response function can provide a good prior distribution for radiance fields, contributing to convergence during training. 

\subsection{Optimization}\label{Sec_Optimization}
As a family of voxel-based neural radiance training methods, we also use the MSE loss and total variational (TV) loss as supervision for optimization.
\begin{equation}
    \mathcal{L} = ||C-\hat{C}||^2 + w_{tv}\mathcal{L}_{tv},
\end{equation}
The $C$ stands for the pixel color of the LDR ground truth image The total variational loss is applied on planes as defined in~\cite{hexplane}. Meanwhile, emptiness-voxel-skipping~\cite{dvgo} and coarse-to-fine training~\cite{hexplane, dvgo} are also employed to regularize and accelerate optimization. For further discussion, please refer to~\cite{hexplane}.


\section{Experiments}

\begin{table}

    \centering

  \begin{tabular}{@{}lccc@{}}
    \toprule
    Model & Accelerated & Dynamic & HDR \\
    \midrule
    NeRF-WT \cite{nerfinthewild} &  & \checkmark&  \\
    HDR-Plenoxels \cite{hdrplenoxels} & \checkmark & &\checkmark \\
    Hexplane \cite{hexplane} & \checkmark & \checkmark &\\
    HDR-NeRF\cite{hdrnerf}&&&\checkmark\\
    Ours & \checkmark & \checkmark &\checkmark \\
    \bottomrule
  \end{tabular}
  \vspace{-10pt}
  \caption{The comparison of different models shows that HDR-HexPlane not only excels in reconstructing dynamic scenes at high speed but also demonstrates the ability to synthesize HDR images. }
 \vspace{-15pt}
 \label{table_model_comparasion}
  
\end{table}

\begin{table*}[ht]  

  \centering
  \resizebox{2\columnwidth}{!}{
  \begin{tabular}{l|ccc|ccc|ccc|ccc}
    \toprule
    \multicolumn{1}{c}{} &\multicolumn{3}{c}{Lego} &\multicolumn{3}{c}{Tank} &\multicolumn{3}{c}{Deer} &\multicolumn{3}{c}{Airplane}\\
    \hline
    Model & PSNR$\uparrow$ & SSIM$\uparrow$ & LPIPS$\downarrow$ & PSNR$\uparrow$ & SSIM$\uparrow$ & LPIPS$\downarrow $& PSNR$\uparrow $& SSIM$\uparrow$ & LPIPS$\downarrow$ &PSNR$\uparrow$ & SSIM$\uparrow $& LPIPS$\downarrow$\\
    \midrule

    HexPlane \cite{hexplane}&15.59 & 0.9130&0.2692& 14.04& 0.4919&0.5794&25.54&0.4025&0.6847&16.66&0.6926&0.3787\\
    NeRF-WT \cite{nerfinthewild}
    &31.55\cellcolor{yellow}&0.9538\cellcolor{yellow}&0.0316\cellcolor{yellow}
    &28.34\cellcolor{yellow}&0.8948\cellcolor{yellow}&0.1615\cellcolor{yellow}
    &28.32\cellcolor{yellow}&0.8676\cellcolor{yellow}&0.1976\cellcolor{yellow}
    &31.92\cellcolor{yellow}&0.9300\cellcolor{yellow}&0.0852\cellcolor{pink}\\
    HDR-Plenoxels \cite{hdrplenoxels}&24.03&0.8103&0.3513&22.52&0.7116&0.5819&24.13&0.7488&0.5043&27.75&0.7779&0.4332\\
    HDR-NeRF\cite{hdrnerf}&26.44&0.8890&0.1650&28.17&0.8520&0.3154&23.74&0.6927&0.5988&26.07&0.9166&0.5041\\
    Ours 
    &36.50\cellcolor{pink}& 0.9786\cellcolor{pink}& 0.0244\cellcolor{pink}
    &33.36\cellcolor{pink}& 0.9184\cellcolor{pink}& 0.1585\cellcolor{pink}
    &31.80\cellcolor{pink}&0.9092\cellcolor{pink}&0.1131\cellcolor{pink}
    &33.88\cellcolor{pink}&0.9316\cellcolor{pink}&0.0997\cellcolor{yellow}\\
    \hline
        \multicolumn{1}{c}{} &\multicolumn{3}{c}{Mutant} &\multicolumn{3}{c}{Punch} &\multicolumn{3}{c}{Standup} &\multicolumn{3}{c}{Jump}\\
    \hline
     & PSNR$\uparrow$ & SSIM$\uparrow $& LPIPS$\downarrow$ & PSNR$\uparrow$ & SSIM$\uparrow $& LPIPS$\downarrow$ & PSNR$\uparrow$ & SSIM$\uparrow$ & LPIPS$\downarrow$ &PSNR$\uparrow$ & SSIM$\uparrow$ & LPIPS$\downarrow$\\
     HexPlane&21.30&0.7176&0.4901&16.50&0.5830&0.4553&20.77&0.8113&0.2457&19.23&0.7290&0.3413\\
     NeRF-WT
     &34.49\cellcolor{pink}&0.9553\cellcolor{pink}&0.0481 \cellcolor{pink}
     &30.32\cellcolor{pink}&0.9316\cellcolor{pink}&0.1029\cellcolor{yellow}
     &31.41\cellcolor{yellow}&0.9348\cellcolor{yellow}&0.0935\cellcolor{pink} 
     &30.46\cellcolor{yellow}&0.9245\cellcolor{yellow}&0.1279\cellcolor{yellow}\\
     HDR-Plenoxels&22.20&0.7889&0.1418&19.95&0.6989&0.3889&28.14&0.8706&0.3281&24.29&0.8015&0.4604\\
          HDR-NeRF&23.09&0.6409&0.6912& 28.16&0.8519 &0.3154& 27.90&0.8941&0.4121&24.52&0.7830&0.4995\\

    Ours
    &33.70\cellcolor{yellow}&0.8879\cellcolor{yellow}&0.1859\cellcolor{yellow}
    &29.89\cellcolor{yellow}&0.9139\cellcolor{yellow}&0.0815\cellcolor{pink}
    &34.84\cellcolor{pink}&0.9472\cellcolor{pink}&0.0949\cellcolor{yellow}
    &33.22\cellcolor{pink}&0.9363\cellcolor{pink}&\cellcolor{pink}0.0815\cr

    \bottomrule
  \end{tabular}
  }
  \vspace{-10pt}
  \caption{Experimental results show that our method, as the first proposed model to handle variation exposures with multi-video input, outperforms other models in most metrics. The \colorbox{pink}{best} and the \colorbox{yellow}{second best} results are denoted by pink and yellow.}\label{table_comparasion_eachscene}
  \vspace{-15pt}
\end{table*}  

\begin{table}[h]
  \resizebox{0.9\columnwidth}{!}{

    \centering
    \begin{tabular}{@{}lcccc@{}}
      \toprule
      Model & PSNR$\uparrow$ & SSIM$\uparrow $& LPIPS $\downarrow$& Training time$\downarrow$ \\
      \midrule
    HexPlane \cite{hexplane} & 18.70 & 0.6424 & 0.4306 & 40 mins \cellcolor{yellow}\\
      NeRF-WT \cite{nerfinthewild} & 30.85 \cellcolor{yellow}& 0.9241\cellcolor{yellow} & 0.1060\cellcolor{pink} & 16 hours \\
      HDR-Plenoxels \cite{hdrplenoxels} & 24.26 & 0.7629 & 0.4539 & 34 mins \cellcolor{pink}\\
      HDR-NeRF\cite{hdrnerf}&26.01&0.8150&0.4377&12 hours\\
      Ours & 33.39\cellcolor{pink}& 0.9278 \cellcolor{pink}& 0.1062 \cellcolor{yellow}& 46 mins \\
      \bottomrule
    \end{tabular}
    }
    \vspace{-10pt}
    \caption{The average metrics across all synthesis datasets. The training time in the table is the average time trained in all scenes.}\label{table_comparasion_total}
    \vspace{-15pt}
\end{table}


In this section, we primarily focus on three main aspects: the evaluation dataset we created, the comparison experiments conducted using different models and the ablation study of our framework. We demonstrate that our model achieves state-of-the-art results on the dataset and perform ablation experiments to evaluate the effectiveness of our proposed method. Additionally, we discuss the limitations of our approach.
\begin{figure*}
  \centering
    \includegraphics[width=\linewidth]{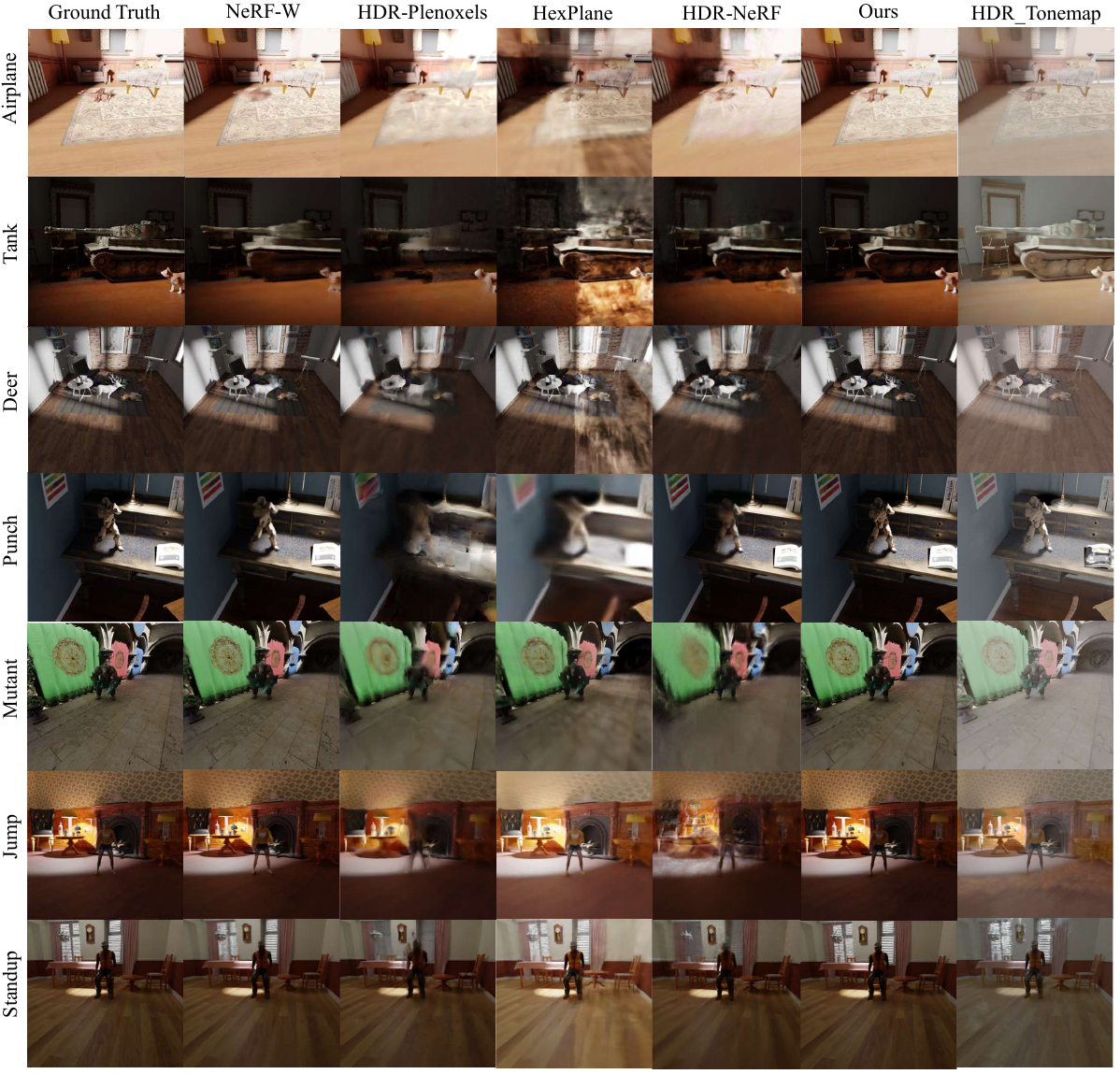}
    \caption{Results of the synthesis datasets with all the images rendered in LDR (Low Dynamic Range). Our method can render dynamic LDR and HDR (High Dynamic Range) images, whereas other methods encountered challenges. The last column shows our tone-mapped HDR image, which balances both overexposed and underexposed areas. }
    \label{fig_comparasion1}
  \hfill
\end{figure*}
\subsection{Experimental Setup}
\paragraph{Datasets.} We build synthetic datasets for evaluation by using Blender to generate eight scenes. The data for these scenes is sourced from~\cite{hdrnerf,blenderswap,mixamo}. Each consists of 80 to 700 images captured from 1$\sim$10 forward-facing cameras, with a resolution of 800$\times$800. The exposures are set between -2EV and 5EV, depending on different scenes. For the Lego, Tank, Deer, and Airplane datasets, their motion primarily represents rigid body movements. We select 80\% of the viewpoint images for the training set and the remaining 20\% for testing. As for the mutant, Standup, Jump, and Punch datasets, they depict rapid non-rigid body movements. We use multiple cameras with varying exposures for capturing these scenes. For training, we select $\frac{6}{7}$ of the viewpoints, while $\frac{1}{7}$ of the viewpoints are reserved for testing. This setup allows us to have varying exposure values for each scene, making it suitable for HDR dynamic scene synthesis.


\paragraph{Training configs. }
We primarily use PyTorch~\cite{pytorch} to implement our model framework. For the configuration of the spatial voxel grid and the positional encoding dimensions for time, coordinates, and directions, as well as the depth of the Multi-Layer Perceptron~\cite{mlp}, we follow the approach used by~\cite{hexplane}.

We keep the learning rate for the time offset parameter at 2e-2 and the learning rate for exposure embedding at 2e-2. The learning rate for the exposure MLP is set to 1e-3. We utilize the Adam optimizer~\cite{adam} as the optimization algorithm with parameters $\alpha$ and $\beta$ set to [0.9, 0.99]. The learning rate is exponentially decayed, reducing to 0.1 times its initial value at the end of training.

For the synthetic dataset, the initial value of the time grid is set to 16 and expanded to 24 during training. We train all the models on a single RTX A5000.

\vspace{-10pt}
\paragraph{Evaluation. }
In the experiment, we compare our proposed method, HDR-HexPlane, with HDR-Plenoxels~\cite{hdrplenoxels}, HexPlane~\cite{hexplane} An enhancement of the NeRF-W model~\cite{nerfinthewild}, referred to
as NeRF-WT~\cite{nerfpl}, has been introduced with transient encoding
capabilities. This enhancement allows NeRF-WT to render dynamic novel views. HDR-NeRF~\cite{hdrnerf} are also considered with given exposure values obtained by Blender~\cite{blender}. ADOP~\cite{adop} relies on point clouds input but colmap~\cite{colmap} cannot provide that because of multi-exposure image input. The main differences between these models are shown in Table~\ref{table_model_comparasion}. Due to the exposure of each image cannot be predicted, we follow the same approach of~\cite{hdrplenoxels,nerfinthewild}, which uses the left half of the images for training and the right half of the image for validation. All the models are trained with officially introduced training configs.

To assess the quality of the synthesized images from new viewpoints, we employ PSNR, SSIM~\cite{ssim}, and LPIPS~\cite{lpips} metrics as evaluation measures. As for comparing HDR images, we usually use the same tone-mapping function as that used in~\cite{hdrnerf}:
\begin{equation}
M(E) = \frac{{\log{ (1 + \mu E)}}}{{\log{(1+ \mu)}}},
\end{equation}

This function provides a mapping from the HDR representation to a standard image format (LDR). The $\mu$ defines the level of compression and is constant to 5000.


\subsection{Results}
In Table~\ref{table_comparasion_eachscene}, we present our experimental results on different datasets and compare the average training time in Table~\ref{table_comparasion_total}. Thanks to the fast dynamic NeRF model and HDR image synthesis pipeline, our method outperforms the respective comparison models across most metrics by a significant margin and accelerates training speed over 10$\times$ compared to NeRF-WT~\cite{nerfpl,nerfinthewild}. While NeRF-WT with transient encoding can learn varying exposures in dynamic scenes, its inability to explicitly model HDR irradiance and reliance on purely implicit MLP representations lead to difficulties in rendering HDR images and slower convergence rates. HDR-Plenoxels~\cite{hdrplenoxels} allows fast learning of scenes but still exhibits considerable blurriness in dynamic scenes while HDR-NeRF~\cite{hdrnerf} failed in reconstructing dynamic scenes. On the other hand, HexPlane~\cite{hexplane}, due to the absence of explicit exposure modeling, tends to overfit the scene exposure learning, achieving promising results on the training set but failing to render high-quality novel view images.

The visualization of the results on the model is displayed in Fig.~\ref{fig_comparasion1}. All scenes contain strong contrasts between light and dark regions, while the dynamic objects in the images are moving at the boundaries of these contrasts. NeRF-WT~\cite{nerfpl,nerfinthewild} performs well in reconstructing the static parts and moderately moving scenes (Mutant, Punch, Jump). However, it exhibits some blurriness when dealing with heavily moving rigid scenes or detailed parts (Deer, Tank, Airplane). HDR-PlenOxels~\cite{hdrplenoxels} can learn the exposure variations in the scenes but lacks explicit dynamic modeling, resulting in blurred rendering of moving objects. On the other hand, HexPlane~\cite{hexplane} shows signs of overfitting in some scenes (Deer, Tank, Airplane). While it achieves good results on the left half of the training images, it exhibits color mixing issues in the right half used for novel view rendering. Through HDR-NeRF~\cite{hdrnerf} can learn the proper CRF, but the lack of modeling dynamic scene and implicit representation results in failure in reconstruction. Additionally, HexPlane~\cite{hexplane} struggles to accurately render the details in some scenes (Punch) or learns the scene colors incorrectly, leading to underfitting for training images with different exposure inputs.
\begin{table}[h]
    \centering
        \resizebox{\columnwidth}{!}{

    \begin{tabular}{@{}lcccc@{}}
      \toprule
      Model & PSNR & SSIM & LPIPS & Training Time \\
      \midrule
      Ours & \textbf{33.39} & \textbf{0.9278} & \textbf{0.1062} & 46 mins \\
      Ours w/ $\phi_c$ & 30.76 & 0.8636 & 0.2295 & 81 mins \\
      Ours w/ $\phi_c$ \& $\mathcal{L}_u$ (Eq.~\ref{eq:zeropointconstraint}) & 30.76& 0.8548 & 0.2315 & 81 mins \\

      Ours w.o/ $\phi_e$  & 32.37& 0.8969 & 0.1881 & \textbf{43 mins} \\
      \bottomrule
    \end{tabular}
    }
\vspace{-10pt}
    
    \caption{
All the results are conducted on all datasets.$\phi_c$ represents replacing the sigmoid CRF with a trainable MLP $\phi_c$ \cite{mlp}. $\mathcal{L}_u$ implies adding the zero-point constraint as referred to in \cite{hdrnerf}. $\phi_e$ stands for the exposure mapping module exposure MLP in Sec.~\ref{learn-exp-ebd}.}
    \label{table_ablation}
\vspace{-15pt}
  
\end{table}

\subsection{Ablation Study}\label{sec_ablation}

\begin{figure}[h]

  \centering
    \includegraphics[width=\linewidth]{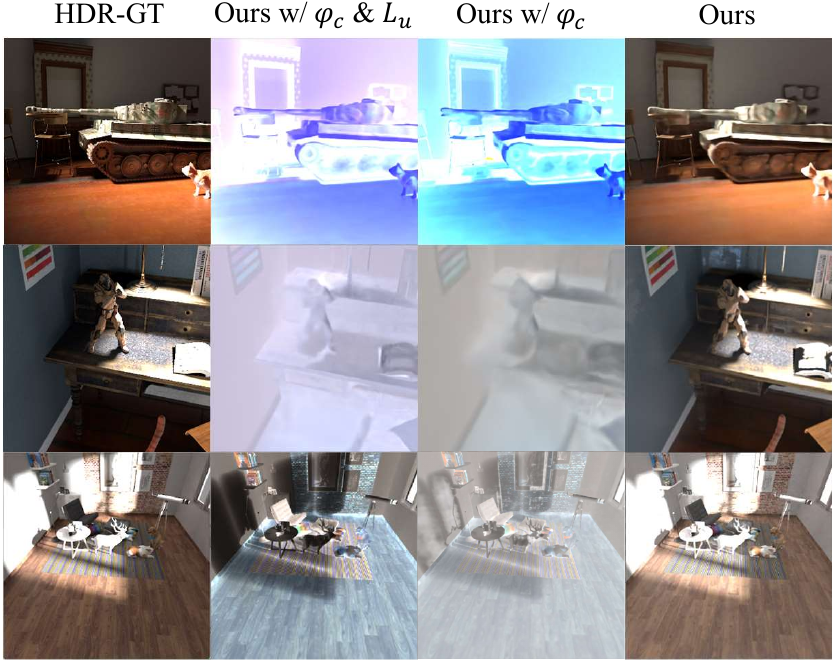}
    \caption{Ablation study of replacing the Sigmoid function by CRF MLP $\phi_c$ like~\cite{hdrnerf}. The first column stands for ground-truth HDR images, the second column stands for replacing CRF $f$ with $\phi_c$ and added by zero-point constraint $\mathcal{L}_u$. the third raw reveals adopting $\phi_c$ only and the last row is our rendered HDR images.}
    \label{fig:ablation}
  \hfill 
\end{figure}
\begin{figure}[h]
  \centering
    \includegraphics[width=\linewidth]{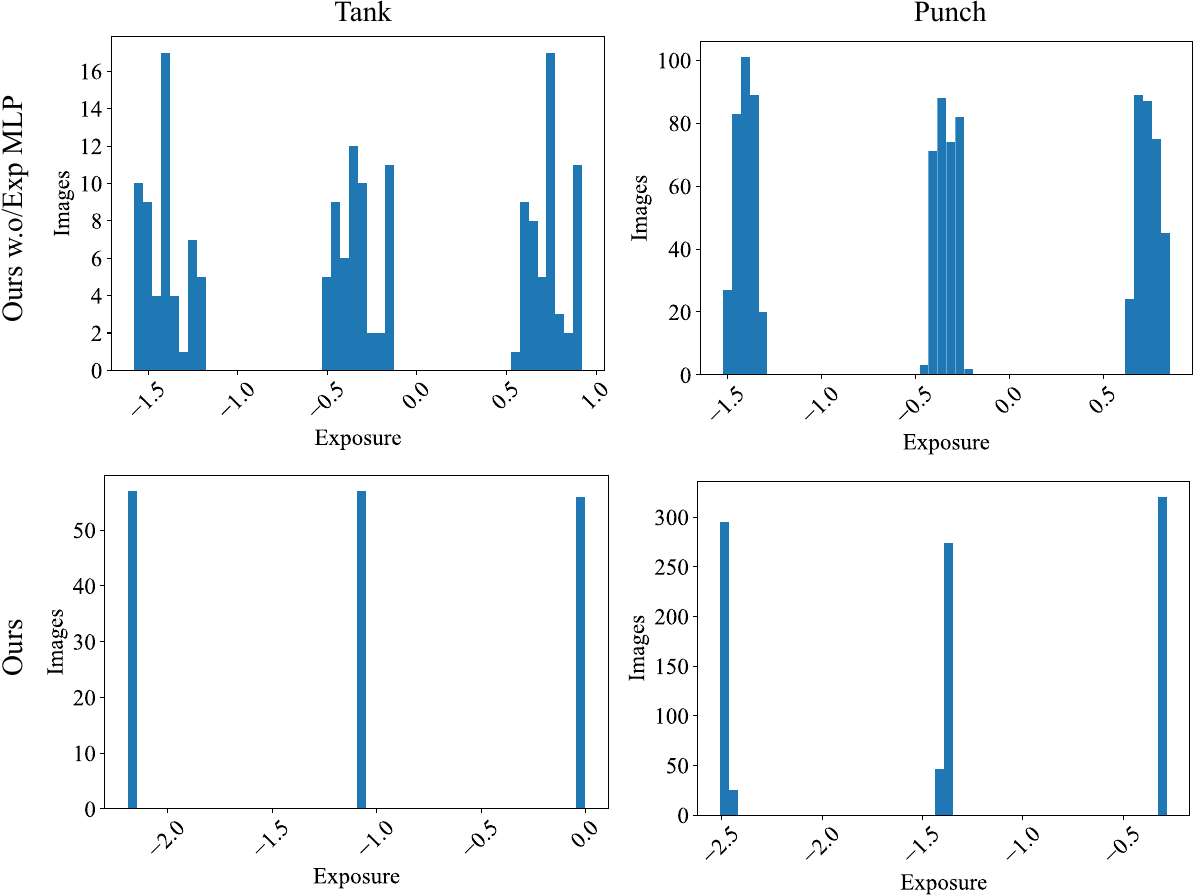}
  \vspace{-10pt}
    
    \caption{The comparison of Learned logarithmic exposure histogram. In the histogram, the horizontal axis represents the logarithmic values of the learned exposure coefficient $e^\prime_j$, and the vertical axis represents the number of images. The first row shows our method without the exposure MLP $\phi_e$, while the second row shows our learned exposure values, which converge to some constant value.}
    \label{fig:hist}
  \hfill
\end{figure}

\paragraph{Camera Response Function.} We evaluate the fixed CRF in comparison with the trainable CRF MLP $\phi_c$ and zero-point constraint $\mathcal{L}_u$ proposed by HDR-NeRF in the second and the third row of Tab.~\ref{table_ablation} and Fig.~\ref{fig:ablation}. Optimizing both CRF and exposure value not only triggers the color bias of rendering HDR images but also causes negative effects on the synthesizing LDR novel views. In HDR-NeRF~\cite{hdrnerf}, CRF can be estimated by known exposure value with zero-point constraint $\mathcal{L}_u$. However, Unknown exposure value triggers confusion in optimizing precise CRF, which leads to color bias in rendering HDR images and lower quality in LDR images. 
\vspace{-10pt}
\paragraph{Exposure Value Mapping.} The ablation of exposure MLP $\phi_e$ is conducted in the last row of Tab.~\ref{table_ablation}. Fig.~\ref{fig:hist} also shows that optimizing them directly leads to the dispersion of learned exposure values. Which also causes a downgrade in rendering quality. Adopting an exposure MLP $\phi_e$ enables HDR-HexPlane to build the correspondence of different input images, make learned exposure value aggregated to the same value.

\vspace{-5pt}
\subsection{Limitations \& Discussion}
The efficient scene reconstruction of NeRF~\cite{nerf} relies heavily on the camera poses computed by COLMAP~\cite{colmap}. However, images captured under different exposures are difficult to match using feature point extraction in COLMAP, leading to errors in camera pose estimation. We notice \cite{rodynrf} may have the potential to facilitate this issue. However, its training speed still needs to be improved, and requires dynamic image masks for training.  

\vspace{-5pt}

\section{Conclusion}
In this paper, we propose HDR-HexPlane, which integrates both HDR imaging and dynamic scene representation pipelines to efficiently learn HDR dynamic scenes. For novel view synthesis, both overexposed and underexposed color regions are considered to achieve state-of-the-art results on the dataset captured by one or more cameras with multi-exposure images. We will reserve the following items for future work, including more robust camera response function modeling and accurate camera pose estimation under multi-exposure settings in complex scenes.
\vspace{-10pt}
\section*{Acknowledgments}
This work was supported by the National Natural Science Foundation of China (No. 62376102). The authors would like to thank Xiaoyu Li for his discussion about details on HDR-NeRF~\cite{hdrnerf}.

{
    \small
    \bibliographystyle{ieeenat_fullname}
    \bibliography{main}
}


\renewcommand\thesection{\Alph{section}}

\setcounter{section}{0}

\begin{figure*}
  \centering
  \vspace{-30pt}
    \includegraphics[width=\linewidth]{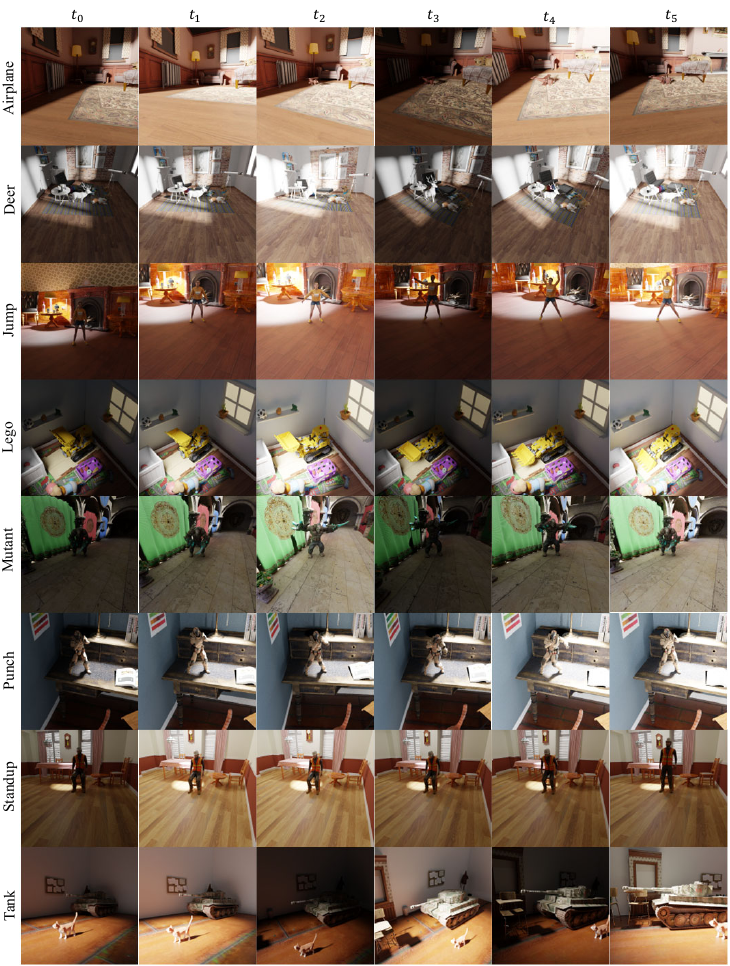}
    \caption{Dataset details for Each Scene. The details for different datasets are outlined below, with $t_0$ through $t_5$ representing relative moments within each dataset. Each row corresponds to distinct motion specifics for various datasets. }
    \label{fig:dataset_details}
  \hfill
\end{figure*}
\begin{figure*}
    \centering
    \includegraphics[width=\linewidth]{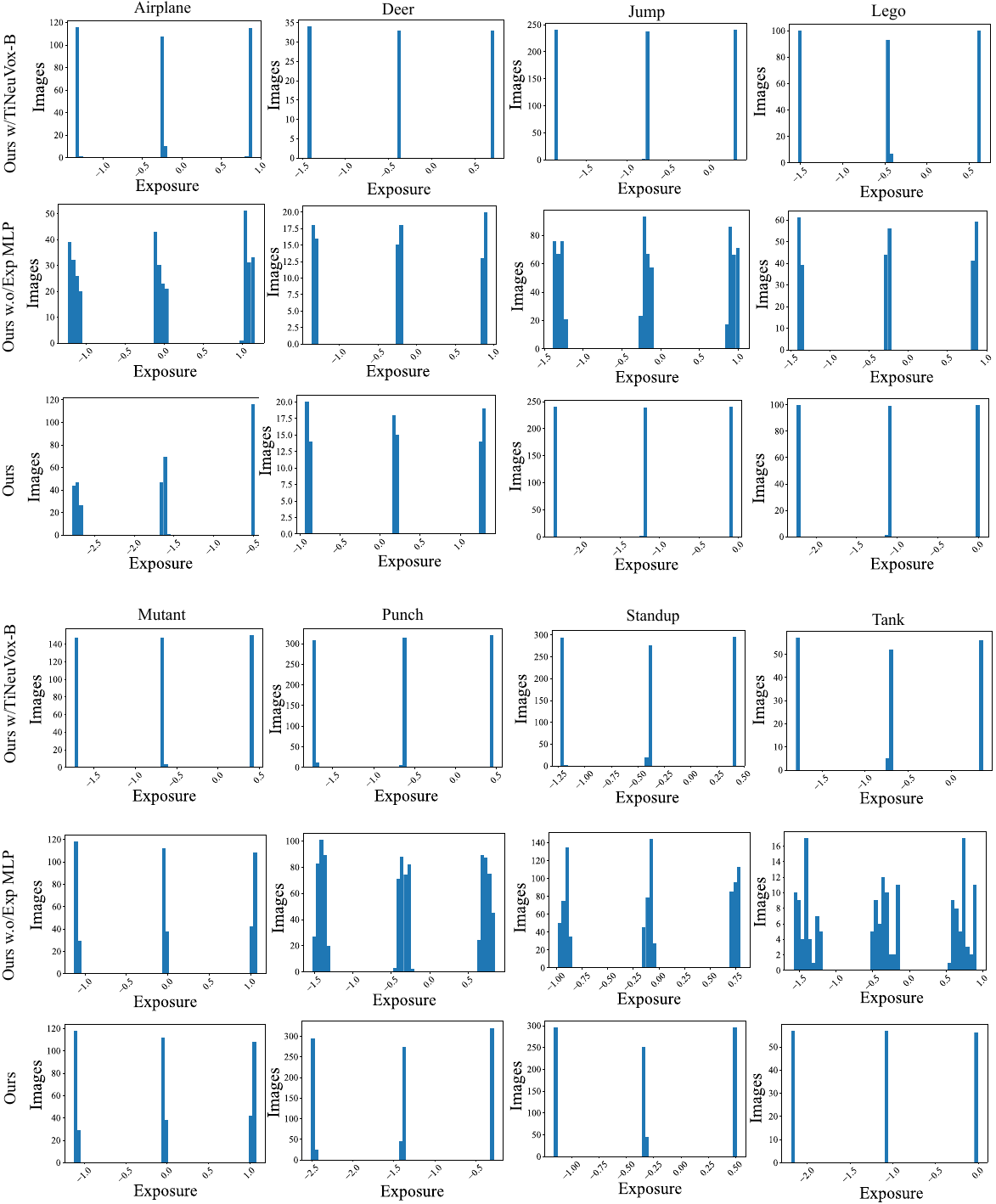}
    \caption{Comparison of exposure values learned across different datasets. The top row suggests that utilizing TiNeuVox\cite{tineuvox} instead of HexPlane\cite{hexplane} as the dynamic scene representation for rendering yields image exposure coefficient distributions. The middle row portrays the outcomes of exposure learning when our model is trained without the Exposure MLP~\cite{mlp} $\phi_e$. The bottom row illustrates the exposure values learned by our model across all datasets. In each chart, the y-axis signifies the count of images at a specific exposure value, while the x-axis corresponds to logarithmic exposure values.}
    \label{fig:image_hist}
  \hfill
\end{figure*}
\begin{figure*}
    \centering
    \includegraphics[width=\linewidth]{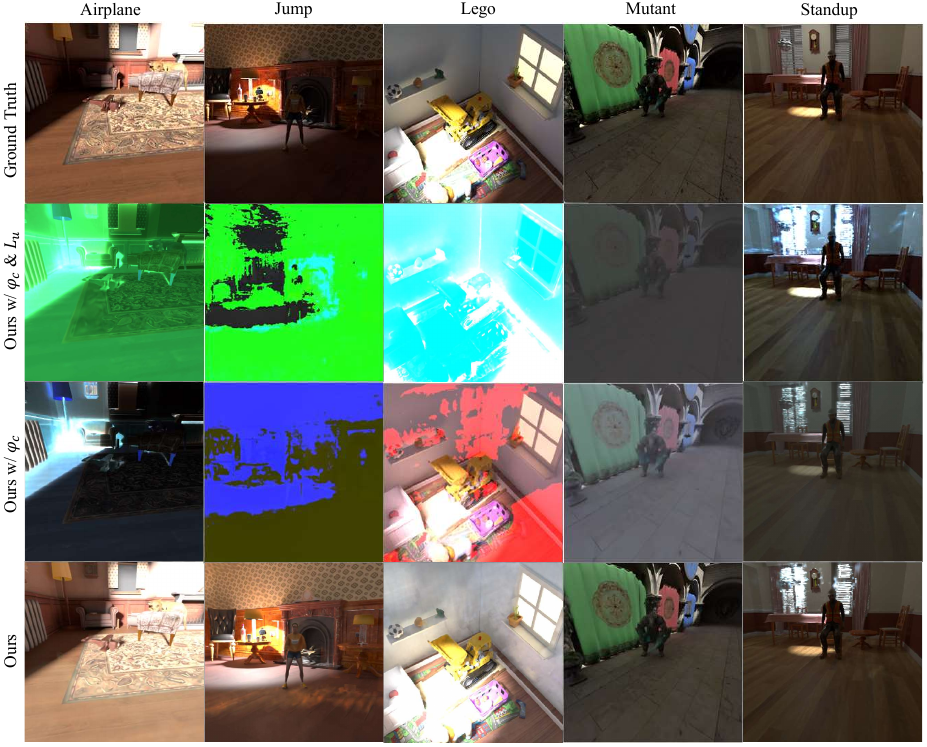}
    \caption{Comparison of results for scenes not presented in the main text. Each column represents a dataset, and each row corresponds to different model configurations or image ground truths.}
    \label{fig:comparasion}
  \hfill
\end{figure*}

\begin{figure}
    \centering
    \includegraphics[width=\linewidth]{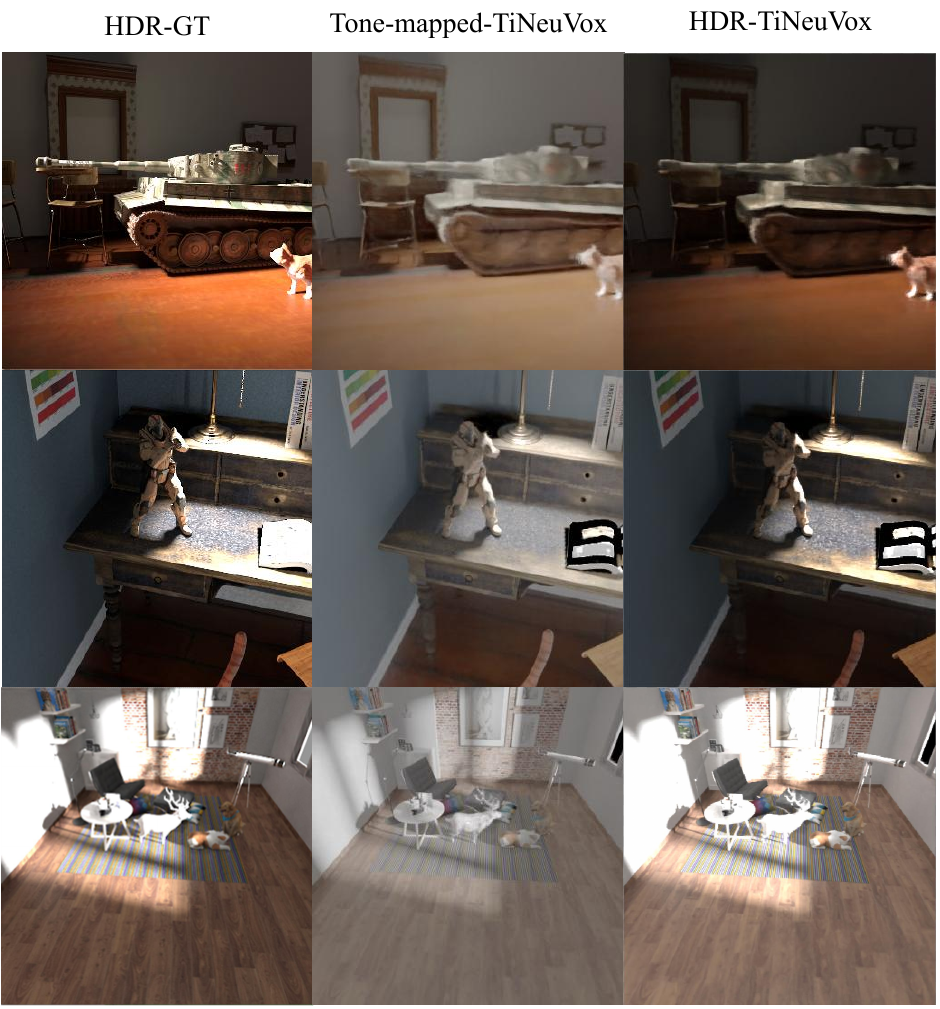}
    \caption{Update the dynamic scene representation for rendering by replacing HexPlane\cite{hexplane} with TiNeuVox-B\cite{tineuvox}, denoted as $f_\beta$. The initial column showcases the actual HDR image, while the second column displays the corresponding HDR image tone-mapped by TiNeuVox-B.}
    \label{fig:tineuvox_run}
  \hfill
\end{figure}

\section{Appendix}

In the supplementary materials, we offer a comprehensive account of the dataset details we created in Sec.~\ref{sec:details of dataset}, coupled with presenting supplementary results from ablation studies in Sec.~\ref{sec:additional ablation studies}. Furthermore, we delve into the topic of replacing our HexPlane representation with other dynamic scene modules in Sec.~\ref{sec:replace by tineuvox}. Finally, more discussions with HDR-NeRF~\cite{hdrnerf} and RawNeRF~\cite{rawnerf} are presented in Sec.~\ref{Sec:morediscussion}.

\subsection{Datasets} \label{sec:details of dataset}
In this section, we primarily elucidate the creation details of the HDR dynamic scene dataset. This includes specifying the number of dynamic frames for each scene and elucidating the process of capturing photographs using cameras within the scene while adhering to the set exposure values. 

We predominantly employ Blender~\cite{blenderswap} for dataset creation. Within each dynamic scene, objects underwent motion, and for every frame during object movement, the camera synchronized its movement while dynamically adjusting exposure for photograph capture. For the creation of the dynamic scene dataset, in the case of rigid body object motion, we find that employing a single camera is sufficient for scene reconstruction. This is attributed to the regular and predictable nature of object motion trajectories. Thus, for datasets such as Lego, Tank, and Airplane we utilized only one camera. While the Deer dataset involves non-rigid body motion, the simplicity of its actions enabled scene reconstruction with only a few frames. The image details of our created datasets are shown in Fig.~\ref{fig:dataset_details}.
\begin{table}
    \centering

  \begin{tabular}{@{}lccc@{}}
    \toprule
    Scene & Motion Frames & Cameras & Exposure Value \\
    \midrule
    Airplane& 350 & 1 & -2,0,2\\
    Deer & 100 & 1 & -2,0,2\\
    Jump & 25 & 10 & 1,3,5 \\
    Lego & 300 & 1 & -3,-1,1 \\
    Mutant & 150 & 3 & -1,1,3 \\
    Punch & 33 & 10 & -1,1,3 \\
    Standup & 60 & 10 & 0,1.5,3\\
    Tank & 170 & 1 & -2,0,2\\

    \bottomrule
  \end{tabular}
  \caption{The dataset setup involves defining the parameters of motion frames, representing the count of frames that capture object movements within a given scene.``Cameras'' refers to the number of cameras utilized to record the scene while objects are in motion. "Exposure Value" denotes the coefficient utilized within the Blender software to regulate exposure settings during the scene capture process.}
 \label{table_dataset_details}
  
\end{table}
In contrast, the Mutant, Standup, Jump, and Punch datasets encompass dynamic scenes with unpredictable actions. Consequently, for each frame, we require multiple perspectives of image data for effective supervision to achieve successful scene reconstruction. Table~\ref{table_dataset_details} presents the detailed configurations adopted for each dataset.
\subsection{Additional Ablation Studies}\label{sec:additional ablation studies}

In this section, we have presented additional results from the ablation study conducted in the main text. These results further underscore the effectiveness and robustness of the model architecture we have designed.

In the comparison of whether or not to include the exposure MLP $\phi_e$ shown in Fig.~\ref{fig:image_hist}, our focus primarily lies in evaluating the learned exposure distribution outcomes. The absence of the $\phi_e$ introduces greater instability in the exposure learning process, thus impacting the synthesis of novel perspectives for LDR images with varying exposures. The impact on image performance metrics is demonstrated in the table within the main text. As observed in Table~\ref{table_dataset_details}, the rendered logarithmic exposure ground truths are equidistant. This property allowed our model to better grasp exposure values. However, since exposure values are initially unknown and the camera response function is fixed as a sigmoid function, which might not align with Blender's camera response function, learned exposure values could exhibit a general bias. Furthermore, upon the removal of the MLP module, we observed inconsistent exposure value distribution learning in the Airplane, Jump, Punch, and Tank datasets, while other datasets (Deer, Lego, Mutant, Standup) showed differing intervals.

We also present additional comparisons in Fig.~\ref{fig:comparasion}, including the use of a CRF MLP $\phi_c$ and the zero-point constraint~\cite{hdrnerf,recoveringhdr} $\mathcal{L}_u$ as supervision. It is evident that, due to the unknown nature of exposure values, joint optimization of exposure and camera response functions without proper priors lead to erroneous learning of HDR images. This results in cases where a color channel's value becomes excessively large, causing color shift deviations in the images.

\subsection{Replace HexPlane Representation by Other Module}\label{sec:replace by tineuvox}
\begin{table}[h]
  \resizebox{\columnwidth}{!}{

    \centering
    \begin{tabular}{@{}lcccc@{}}
      \toprule
      Representation & PSNR$\uparrow$ & SSIM$\uparrow $& LPIPS$ \downarrow$& Training time$\downarrow $\\
      \midrule

      TiNeuVox-B~\cite{tineuvox}& \textbf{33.94}& 0.9241 & 0.1956 & 50 mins \\
      HexPlane~\cite{hexplane} &33.39 & \textbf{0.9279}& \textbf{0.1061} & \textbf{46 mins}\\
      \bottomrule
    \end{tabular}
    }
    \vspace{-10pt}
    \caption{Use TiNeuVox-B instead of our HexPlane~\cite{hexplane} representation and get the average of the indicators in our provided 8 scenes.}\label{table:Tineuvox-results}
     \end{table}
The approach we introduced can be employed not only with HexPlane~\cite{hexplane} but also with various other dynamic scene representations. In this section, we substitute the HexPlane representation $f_\beta$ with TiNeuVox~\cite{tineuvox}. The modifications made to TiNeuVox follows the same principle as those for HexPlane. Our experimental outcomes are summarized in Table~\ref{table:Tineuvox-results}, indicating that utilizing TiNeuVox as a foundational model for dynamic scene rendering yields favorable results on our datasets as well. Fig.~\ref{fig:image_hist} and Fig.~\ref{fig:tineuvox_run} display the novel views images and exposure coefficients using TiNeuVox as the underlying architecture $f_\beta$. Therefore, for the reconstruction of HDR dynamic scenes, our proposed technique can be adapted to various dynamic scene models as a baseline. For more in-depth discussions about HexPlane and TiNeuVox, kindly refer to~\cite{hexplane}.

\subsection{More Discussions}
\label{Sec:morediscussion}
\paragraph{HDR-NeRF.} HDR-NeRF~\cite{hdrnerf} is an elegant vanilla NeRF-based method that utilizes an MLP and known camera exposure value to estimate CRF. However, the collection of camera exposures will be needed in additional steps and the estimated CRF is dependent on different cameras. We propose a unified pipeline that fixes the curve of CRF and enables the model to learn the exposure value of different camera inputs. Besides, the lack of modeling dynamic scenes in \cite{hdrnerf} also causes the failure to handle more complex cases. 

\paragraph{RawNeRF.}
RawNeRF~\cite{rawnerf} can also reconstruct static HDR radiance fields with raw image input. However, images in raw format need to be purposefully captured to create a specific dataset for reconstructing the HDR neural radiance field. Plain LDR images, used in our paper, under different lighting conditions can be easily obtained in the vast majority of scenes, offering a broader potential for application.


\end{document}